\documentclass[letterpaper, 10 pt, conference]{ieeeconf}  

\usepackage{graphicx}
\usepackage{amsmath}
\usepackage{amssymb}
\usepackage{stfloats}

\usepackage{xcolor}
\usepackage{url}
\usepackage{tensor}
\usepackage{mathtools}
\usepackage{amsmath}
\usepackage{bbm}
\usepackage{enumerate}
\usepackage[ruled,linesnumbered]{algorithm2e}
\usepackage{siunitx}
\usepackage{tabularx}
\usepackage{multirow}
\newcommand{\tabincell}[2]{\begin{tabular}{@{}#1@{}}#2\end{tabular}}

\DeclareMathOperator{\updateProbePose}{updateProbePose}
\DeclareMathOperator{\sample}{sample}
\DeclareMathOperator{\confMap}{confMap}

\IEEEoverridecommandlockouts                              

\title{\LARGE \bf
Autonomous Navigation of an Ultrasound Probe Towards Standard Scan Planes with Deep Reinforcement Learning
}

\author{Keyu Li, Jian Wang, Yangxin Xu, Hao Qin, Dongsheng Liu, Li Liu$^{*}$, and Max Q.-H. Meng$^{*}$, \textit{Fellow}, \textit{IEEE}
\thanks{This work was partially supported by National Key R\&D program of China with Grant No. 2019YFB1312400, Hong Kong RGC GRF grant \#14210117, Hong Kong RGC TRS grant T42-409/18-R  and Hong Kong RGC GRF grant \#14211420 awarded to Max Q.-H. Meng.}
\thanks{K. Li, Y. Xu, and L. Liu are with the Department of Electronic Engineering, The Chinese University of Hong Kong, Hong Kong, China (e-mail: kyli@link.cuhk.edu.hk; yxxu@link.cuhk.edu.hk; liliu@cuhk.edu.hk).}%
\thanks{J. Wang is with the School of Biomedical
Engineering, Shenzhen University, Shenzhen, China.}
\thanks{H. Qin is with Sonoscape Medical Corp., Shenzhen, China.}
\thanks{D. Liu is with the Department of Pain, Peking University Shenzhen Hospital, Shenzhen, China.}
\thanks{Max Q.-H. Meng is with the Department of Electronic and Electrical Engineering of the Southern University of Science and Technology in Shenzhen, China, on leave from the Department of Electronic Engineering, The Chinese University of Hong Kong, Hong Kong, and also with the Shenzhen Research Institute of the Chinese University of Hong Kong in Shenzhen, China (e-mail: max.meng@ieee.org).}%
\thanks{$^{*}$Corresponding authors.}%
}

\begin{document}

\maketitle
\thispagestyle{empty}
\pagestyle{empty}

\begin{abstract}
Autonomous ultrasound (US) acquisition is an important yet challenging task, as it involves interpretation of the highly complex and variable images and their spatial relationships. 
In this work, we propose a deep reinforcement learning  framework to autonomously control the 6-D pose of a virtual US probe based on real-time image feedback to navigate towards the standard scan planes under the restrictions in real-world US scans. Furthermore, we propose a confidence-based approach to encode the optimization of image quality in the learning process.
We validate our method in a simulation environment built with real-world data collected in the US imaging of the spine. Experimental results demonstrate that our method can perform reproducible US probe navigation towards the standard scan plane with an accuracy of $4.91mm/4.65^\circ$ in the intra-patient setting, and accomplish the task in the intra- and inter-patient settings with a success rate of $92\%$ and $46\%$, respectively. The results also show that the introduction of image quality optimization in our method can effectively improve the navigation performance.

\end{abstract}
\section{INTRODUCTION}

Due to the advantages of portability, non-invasiveness, low cost and real-time capabilities over other imaging techniques, ultrasound (US) imaging has been widely accepted as both a diagnostic and a therapeutic tool in various medical disciplines \cite{ultrasound}. However, US acquisition in current clinical procedures requires specialized personnel to manually navigate the probe towards the correct imaging plane, which is very time-consuming and the imaging quality is highly dependent on the sonographer. Moreover, the heavy workload has been exposing the sonographers to health risks such as work-related musculoskeletal disorders \cite{disorder1}\cite{disorder2}. Therefore, an automation of the US scanning process holds great promise for reducing the heavy workload of sonographers, shortening the examination time, yielding high-quality, standardized and operator-independent imaging results, and improving access to care in remote or rural communities.

\begin{figure}[t]
\setlength{\abovecaptionskip}{0.1cm}  
\centering
\includegraphics[scale=1.0,angle=0,width=0.48\textwidth]{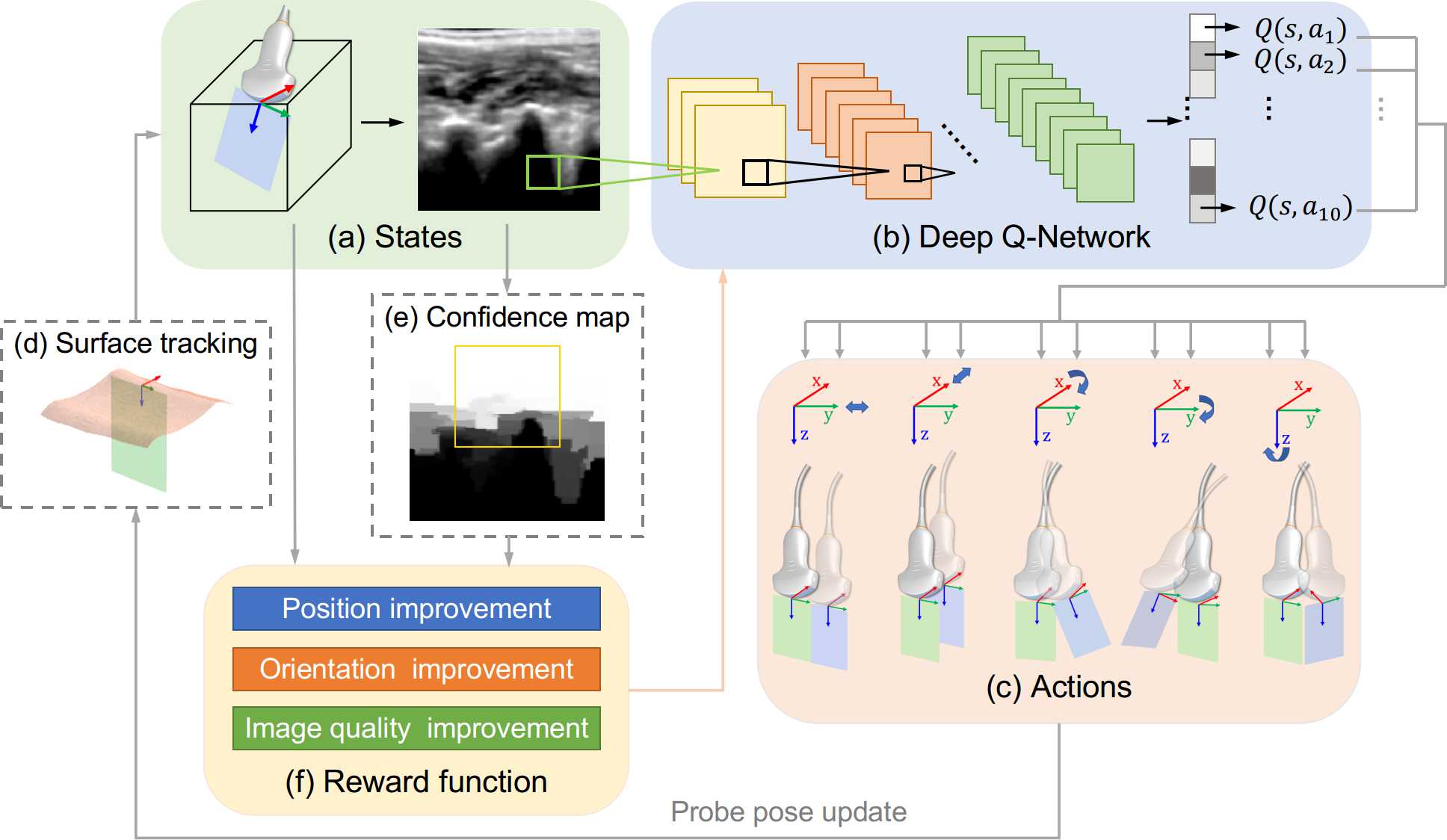}
\caption{An overview of the presented method for autonomous navigation of a US probe. At each time step, a 2D image is acquired with the current probe pose (a) and serves as the input of the deep Q-network (b). The optimal movement action is selected from 10 actions associated with the 5-DOF pose of the probe (c), and 1 DOF is used to track the patient surface (d). The confidence map (e) is computed from the US image to estimate the image quality, and the reward function (f) encourages the improvement of position, orientation and image quality during the navigation.}
\label{Fig_method_overview}
\vspace{-0.5cm}  
\end{figure}

Over the past few decades, the advances in robotic US acquisitions have demonstrated the potential of using robots to automatically cover a region of interest in the patient \cite{mustafa2013robio}\cite{Hennersperger2016MRI}\cite{virga2016iros}\cite{huang2018robotic}. 
However, these methods ignore the interpretation of the images and the precise positioning of the probe for visualization of the \textit{standard scan planes}, which are important imaging planes in clinical US examinations that can clearly show the anatomical structures of the target tissue and contain valuable information for identifying abnormalities or performimg biometric measurements \cite{baumgartner2017sononet}\cite{spineUS}. Autonomous probe navigation towards the standard scan planes remains a challenging task because it involves interpretation of the highly complex and variable US images acquired during the scan and their spatial relationships.

In recent years, Reinforcement Learning (RL) has achieved superior performance in sequential decision-making problems in many real-life applications such as mobile robot navigation \cite{li2019sarl}. Inspired by the latest developments in RL-based medical image analysis \cite{alansary2018automatic}\cite{alansary2019evaluating}, in this work, we propose a deep RL framework for the autonomous US acquisition task, where an agent continuously controls the 6-D pose of a virtual US probe based on the acquired images to navigate towards the standard scan plane, as shown in Fig. \ref{Fig_method_overview}. Moreover, we encode the optimization of image quality in the RL framework with a confidence-based approach to improve the navigation performance. The video demonstration can be found at \url{https://youtu.be/_jcmyAA0WxM}. The main contributions of this work are as follows:
\begin{itemize}
\item We present the first deep RL framework to control the 6-D pose of a virtual US probe based on real-time US image feedback, in order to mimic the decision-making process of sonographers to autonomously navigate towards the standard scan planes. 
\item Furthermore, based on the \textit{ultrasound confidence map} \cite{confimap}, we propose a confidence-based method to introduce the optimization of image quality in the learning of navigation strategy. We empirically demonstrate in the experiments that our method can improve the navigation performance. 
\item A simulation environment is delicately designed and built with real-world US data to simulate the US probe navigation scenario. We take into account several practical requirements in real-world US scans to make the learned navigation policy easier to generalize to real-world applications.
\end{itemize}

%

\section{RELATED WORK}
A number of methods have been proposed to detect the standard scan planes from 3D US volumes. Some methods used CNNs to regress the transformation from the current plane to the standard plane \cite{schmidt2019offset}\cite{li2018standard}, but this kind of prediction may cause abrupt changes in position rather than gradual and continuous changes, which is undesirable for the probe navigation task. Alansary et al. \cite{alansary2018automatic} parameterized a plane as $ax+by+cz+d=0$, and customized an RL agent to learn step-by-step adjustment of the plane parameters to find the standard plane in 3D MRI scans. Dou et al. \cite{dou2019agent} extended this method for US standard plane detection. However, these methods focus on detecting the planes in the pre-acquired and processed 3D volumes and cannot be directly used to navigate a probe.

Some researchers addressed the probe navigation problem with imitation learning methods such as inverse reinforcement learning \cite{burke2020learning} and behavioral cloning \cite{droste2020automatic} to learn from expert demonstrations. However, complete and accurate expert demonstrations can be intractable or expensive to obtain in the clinical US scans. Jarosik et al. \cite{jarosik2019automatic} customized an RL agent to move a virtual probe in a simple and static toy environment, but the real-world probe navigation task is much more complicated and challenging due to the highly variable anatomy among patients.
In \cite{milletari2019straight}, the researchers used RL to learn cardiac US probe navigation in a simulation environment built with spatially tracked US frames acquired by a sonographer on a grid covering the chest of the patient. Similarly, \cite{hase2020ultrasound} used 2D images acquired on a grid covering the lower back to train an RL agent to find the sacrum with 2-DOF actions. Since the simulation environments in these grid-based methods are built with 2D US images acquired by a limited number of probe poses, the actions learned by these methods are restricted to the collected data. Also, the pose of the patient relative to the probe is assumed to be static, which is difficult to achieve in real US scanning scenarios. 

In addition, prior works \cite{burke2020learning}--\cite{hase2020ultrasound} only controls a part of the probe pose based on the image feedback instead of the complete 6-DOF probe pose, and none of these methods consider the influence of image quality on the navigation performance. In this work, we build a simulation environment using 3D US volumes reconstructed from real-world US data covering the region of interest in the patient. Therefore, the agent can arbitarily change the 6-DOF pose of the virtual probe and the resulting US image can be sampled in the volume. Besides, we adopt a probe-centric action parameterization to relax the requirements for the patient's pose during the scan, thereby making the learned probe navigation policy easier to generalize to real-world applications. Moreover, we propose a method to take into account the image quality optimization in the learning of navigation strategy.

\section{METHODS}

\subsection{Reinforcement Learning for Probe Navigation}

In this work, we consider the autonomous US acquisition task where a US probe is automatically navigated to obtain an image of the standard scan plane. This problem can be formulated as a sequential decision-making problem in the RL framework, where the agent, in this case a virtual US probe, interacts with the environment, in this case the virtual patient, in a sequence of observations, actions and rewards. We define the RL framework for probe navigation as follows.

\subsubsection{States}
In our environment, the virtual patient is a reconstructed 3D US volume $V$ that covers the region of interest in the patient. We set the virtual US probe as a commonly used 2D probe with a field-of-view of $h \times w$. At time step $t$, the 6-D pose of the probe $\{P\}$ with respect to the world coordinate frame $\{W\}$ can be described by a spatial transformation matrix $\prescript{W}{P}{T}_t$, which uniquely determines the probe's position $\boldsymbol{p}_t=[p_x,p_y,p_z]$ and orientation $\boldsymbol{q}_t=[q_x,q_y,q_z,q_w]$ (represented with a quaternion). As shown in Fig. \ref{Fig_method_overview} (a), assuming that the $yz$ plane of the probe is aligned with the image plane, then the 2D US image of size $h\times w$ acquired with the current probe pose can be sampled from the patient $I_t \leftarrow sample(V, \boldsymbol{p}_{t}, \boldsymbol{q}_{t})$. The goal probe pose is $(\boldsymbol{p}_g$, $\boldsymbol{q}_g)$, corresponding to the standard plane image $I_g\leftarrow sample(V, \boldsymbol{p}_{g}, \boldsymbol{q}_{g})$. We consider the visual navigation task as patially observed, where the goal and the probe pose in the world coordinate system are unobservable, and the agent can only observe the acquired US images. A sequence of $m$ recent images are stacked together as the state $s_t:=[I_{t-m+1},\cdots, I_{t}]$ to take into account the dynamic information  \cite{mnih2015human}.

\subsubsection{Actions}
Based on the observation at time step $t$, the agent takes an action selected by its policy $\pi\colon s_t \mapsto a_t$. Basically, we define the navigation action as a transform operator in the probe frame $a_t:= \prescript{P}{}{T}_t$ that transforms the current probe pose to a new pose as
\begin{equation}
\vspace{-0.1cm}
\prescript{W}{P}{T}_{t+1} = \prescript{W}{P}{T}_t \cdot \prescript{P}{}{T}_t
\label{basic_pose_update}
\end{equation}

Different from \cite{milletari2019straight}\cite{hase2020ultrasound} which represent the actions in the world coordinate frame, we adopt a probe-centric action parameterization to relax the restrictions on the patient's actual pose in the world. 
We only require the coronal plane of the patient to be roughly parallel to the horizontal plane ($xy$ plane of $\{W\}$). As shown in Fig. \ref{Fig_method_overview} (c), $10$ discrete actions related with $5$ DOFs of the probe are used, namely, $4$ actions to translate a certain distance $\pm d_{step}$ along the probe's $x, y$ axes and $6$ actions to rotate a certain angle $\pm \theta_{step}$ around the probe's $x, y, z$ axes, respectively. Since we use the height of the probe to track the patient surface (which will be explained in \textit{3)}), we slightly modify the $4$ translational movement actions to translation along the projections of the probe's $x,y$ axes on the horizontal plane $x',y'$.

Similar to \cite{alansary2018automatic}\cite{alansary2019evaluating}, we use hierarchical action steps to search for the plane in a coarse-to-fine manner. Specifically, a total of $5$ step sizes are used. The action step is initialized as $d_{step}=5mm$ and $\theta_{step}=5^\circ$, and a buffer is used to store $30$ historical poses $[(\boldsymbol{p}_{t-29},\boldsymbol{q}_{t-29}), \cdots,(\boldsymbol{p}_{t},\boldsymbol{q}_{t})] $. If $3$ pairwise Euclidean distances between the historical poses are less than a threshold $0.01$, the agent is considered to have converged to a pose and the action step will be reduced by $1$ unit until it becomes zero. 

\subsubsection{State transition under restrictions}
If there are no restrictions, the probe pose can be updated according to the selected action as in the previous work \cite{jarosik2019automatic}\cite{milletari2019straight}\cite{hase2020ultrasound}. Here, we instead consider two requirements for the probe pose in real-world US scans: i) the contact between the probe and the patient surface should be maintained to ensure sufficient acoustic coupling, and ii) the tilt angle of the probe should be limited to ensure the comfort and safety of the patient. 

To this end, we first update the horizontal position of the probe ($p_x,p_y$) using (\ref{basic_pose_update}), and use the $z$-coordinate of the probe to track the patient surface $p_z \leftarrow surface(p_x,p_y)$, as shown in Fig. \ref{Fig_method_overview} (d). In order to extract the surface equation $z=surface(x,y)$, for each pair of ($x,y$) in the volume $V$, we approximate the surface point as the point with the largest $z$-coordinate whose gray value is not zero. Note that this intensity-based method is only used to estimate the patient surface in our simulation. In real-world applications, the patient surface can be extracted in real time based on data obtained with external sensors such as an RGB-D camera \cite{Hennersperger2016MRI}\cite{virga2016iros}. 
Second, after the new probe orientation is given by (\ref{basic_pose_update}), the resulting tilt angle of the probe (i.e., angle between the $z$-axis of the probe $\hat{\boldsymbol{z}}_{p}$ and the $-z$ direction of $\{W\}$) is $\beta=\arccos (\hat{\boldsymbol{z}}_p,[0,0,-1]^{T})=\arccos (-\prescript{W}{P}{T}_{t+1}(3,3))$. We restrict the tilt angle to be smaller than $30^\circ$. If $\beta > 30^\circ$, the probe orientation will not be updated. After the new probe pose $\boldsymbol{p}_{t+1}, \boldsymbol{q}_{t+1}$ is obtained under the above restrictions, a new US image $I_{t+1}$ can be acquired, and the observation is updated to $s_{t+1}$.

\subsubsection{Reward function}
In our probe navigation task, the reward function should be designed to encourage the agent to move towards the goal. Instead of simply classifying the results of actions as moving closer to or further away from the goal and assigning corresponding rewards \cite{alansary2018automatic}\cite{milletari2019straight}\cite{hase2020ultrasound}, we design the reward function to be proportional to the amount of pose improvement. At time step $t$, the distance-to-goal can be measured in position and orientation, respectively:
\begin{equation} 
\label{F_rate}
d_t = \left\|\boldsymbol{p}_t-\boldsymbol{p}_g\right\|_2, \ 
\theta_t = 2 \arccos (| \langle \boldsymbol{q}_t, \boldsymbol{q}_g \rangle |),
\end{equation}
where $d_t$ is the Euclidean distance between the current positions of the probe and the goal, and $\theta_t$ is the minimum angle required to rotate from the current probe orientation to the goal orientation. Then, the pose improvement after taking action $a_t$ normalized with the step size is
\begin{equation} 
\label{F_rate}
{\Delta d}_t = \frac{d_{t} - d_{t+1}}{d_{step}},\ 
{\Delta \theta}_t = \frac{\theta_{t} - \theta_{t+1}}{\theta_{step}},
\end{equation}
where ${\Delta d}_t,{\Delta \theta}_t \in [-1,1]$.

In addition, we assign a high reward ($+10$) for task  accomplishment ($d_t \leq 1 mm \  \text{and } \theta_t \leq 1 ^\circ$) and add some penalties based on the restrictions of the environment. If the action causes the tilt angle of the probe $\beta>30^\circ$, the agent will receive a penalty of $-0.5$. If the probe moves outside the patient (the proportion of pixels with non-zero gray value in $I_t$ is less than $30\%$), the agent will get a penalty of $-1$. In summary, the reward function (without confidence improvement) is defined as
\begin{equation} 
\label{reward}
r_t =
  \begin{cases}
  -1, & \text{if moving out of } V ;\\
  -0.5, & \text{if } \beta > 30^\circ;\\
  10, & \text{if reaching goal} ; \\
  {\Delta d}_t+{\Delta \theta}_t, & \text{otherwise}. \\
  \end{cases}
\end{equation}

\subsubsection{Termination conditions}
During training, we terminate an episode when: a) the goal is reached, or b) the number of steps exceeds the maximum limit, or c) the action step is reduced to zero, or d) the probe moves out of the patient. During testing, only the termination conditions b,c,d are used due to the absence of the goal's true location.

\begin{algorithm} \small 
\caption{\small SonoRL Training}
\label{algorithm3}
\KwIn{patient dataset $D$}
\KwOut{$Q$-network}
Initialize experience replay memory $E$ with demonstration\;
Pre-train the network $Q$ with $E$\;
Initialize target network $\widehat{Q} \leftarrow Q$\;
Initialize steps $n=0$\;
\While{steps $n < N $}{
Sample a patient US volume $V$ and goal probe pose $\boldsymbol{p}_{g},\boldsymbol{q}_{g}$ from $D$\;
Extract patient surface $z=surface(x,y)$\;
Initialize time $t=0$, action step $d_{step},\theta_{step}$\; Initialize probe pose 
$\boldsymbol{p}_{t},\boldsymbol{q}_{t}$ randomly, observe an image $I_{t} \leftarrow \sample(V,\boldsymbol{p}_{t}, \boldsymbol{q}_{t})$ and initialize observation $s_{t}$\;
Calculate ROI confidence $c_{t}$\;
\Repeat{Termination condition satisfied}{
Select a random action $a_t$ with probability $\varepsilon$, otherwise select $a_t = \arg \max \limits_{a} Q(s_t,a)$\;
Update probe pose under restrictions
$\boldsymbol{p}_{t+1},\boldsymbol{q}_{t+1} \leftarrow \updateProbePose(\boldsymbol{p}_{t},\boldsymbol{q}_{t},a_t,surface)$\;
Observe new image $I_{t+1} \leftarrow \sample(V,\boldsymbol{p}_{t+1}, \boldsymbol{q}_{t+1})$ and update observation $s_{t+1}$ \;
Calculate ROI confidence $c_{t}$\;
Calculate reward $r_{t}$ by (\ref{reward}) or (\ref{reward_with_conf})\;
Store transition $(s_t,a_t,r_t,s_{t+1})$ in $E$\;
Update action step $d_{step},\theta_{step}$ if needed\;
$t \leftarrow t+1$\;
$n \leftarrow n+1$\;
\If{($n \mod K)=0$}{
Sample a minibatch data from $E$\;
Update network $Q$ by gradient descent\;}
\If{($n \mod C)=0$}{
Update target network $\widehat{Q} \leftarrow Q$\;}
}
}
\Return{$Q$}
\vspace{-0.9mm}
\end{algorithm}

\subsection{Confidence-aware Agent}
In clinical US examinations, the sonographer will continuously adjust the probe to obtain clear images while searching for the correct imaging plane, and avoid positions that may cause poor image quality. This motivates us to take into consideration the impact of image quality on the agent's navigation performance. Similar to \cite{virga2016iros}\cite{2017confidriven}, we evaluate the image quality using the \textit{ultrasound confidence map} \cite{confimap}, which estimates the pixel-wise confidence in the image based on a random
walks framework, as shown in Fig. \ref{Fig_method_overview} (e). At time step $t$, the confidence map $C_t\leftarrow \confMap(I_t)$ is computed from the US image, $C_t(i,j)\in [0,1]$. Let $S$ denote the region of interest (ROI) in the image, the quality of the image $I_t$ can be represented by the average ROI confidence
\begin{equation}  
c_t=\frac{1}{|S|}\sum_{(i,j)\in S}{C_t(i,j)}
\label{eq_conf}
\end{equation}

The improvement of image quality after taking action $a_t$ can be represented by ${\Delta c}_t = c_{t+1} - c_{t}$. We hypothesize that encouraging the improvement of image quality in addition to reducing the distance-to-goal can help the agent learn a better navigation strategy. This has been empirically verified by the experimental results in Section~IV-C. Therefore, we introduce a confidence-aware auxiliary term in the reward function to encode the optimization of image quality in the learning process. As shown in Fig. \ref{Fig_method_overview} (f), the modified reward function for the confidence-aware agent takes into consideration the improvement of position, orientation and image quality:
\begin{equation} 
r_t =
  \begin{cases}
  -1, & \text{if moving out of } V ;\\
  -0.5, & \text{if } \beta > 30^\circ;\\
  10, & \text{if reaching goal} ; \\
  {\Delta d}_t+{\Delta \theta}_t+{\Delta c}_t, & \text{otherwise}. \\
  \end{cases}
\label{reward_with_conf}
\end{equation}

\subsection{Deep Q-Network Training}
\subsubsection{Deep Q-Learning Algorithm}
In the RL framework, the agent learns to maximize the discounted sum of future rewards $G_t=r_{t+1}+\gamma r_{t+2}+\gamma^{2} r_{t+3}+\cdots + \gamma^{T-t-1}r_T$ where $\gamma\in(0,1)$ is a discount factor and $T$ is the time step when the episode is terminated. The optimal policy  $\pi^{*}\colon s_t \mapsto a_t$ is to maximize the  expected return\vspace{-0.3cm}

\begin{equation}
\begin{aligned}
\pi^*&=\arg \max \limits_{a}Q^{*}(s,a),\\
Q^{*}(s,a)&=\mathbb{E}_{s'}[r(s,a)+\gamma \max \limits_{a'} Q^{*}(s',a')]
\end{aligned}
\end{equation}
where the optimal state-action value function $Q^{*}(s,a)$ is defined as the maximum expected return following any policy $\pi$: $Q^{*}(s,a)=\max \limits_{\pi} \mathbb{E}_{\pi}[G_t|s_t=s,a_t=a]$.

The deep Q-learning algorithm \cite{mnih2015human} uses a deep neural network to approximate the Q-function, and train the network by the temporal-difference method with experience replay and target network techniques. We implement the deep Q-learning algorithm with minor modifications targeted at our application. We refer to our DQN-based RL framework for US probe navigation as \textit{SonoRL}, which is outlined in Algorithm 1. The Q-network is pre-trained on some demonstration trajectories generated with an expert policy which selects actions to maximize the one-step pose improvement ${\Delta d+\Delta \theta}$ (line 1-3). Subsequently, the network is trained with self-generated experiences by interacting with the environment with an $\varepsilon$-greedy policy (line 4-29). Two different agents are used, i.e., \textit{SonoRL w/ conf} and \textit{SonoRL w/o conf}, which use different reward functions (line 16).

\subsubsection{Implementation Details}
We adopt the SonoNet-16 \cite{baumgartner2017sononet} architecture initially proposed for US standard plane detection as our Q-network model and remove the final softmax layer (see Fig. \ref{Fig_method_overview} (b)). The network is trained every $10$ interaction steps with a batch size of $32$ using Adam optimizer \cite{kingma2014adam}, and the target nerwork is updated every $1k$ training steps. The discount factor $\gamma$ is $0.9$. The exploration rate $\varepsilon$ decays linearly from $0.5$ to $0.1$ in the first $100k$ interaction steps and stays unchanged for the remaining steps. The experience replay memory has a capacity of $100k$ and is initialized with $5k$ demonstration data. During the pre-training phase, $100k$ demonstration experiences are collected and the network is updated for $10k$ steps with a learning rate of $0.01$. During reinforcement learning, the learning rate is set to $0.01$ for the first $40k$ training steps, $0.001$ for the next $40k$ steps, $5\text{e-}4$ for the next $30k$ steps, and $1\text{e-}4$ for the remaining steps.

\begin{figure}[bt]
\centering
\includegraphics[scale=1.0,angle=0,width=0.48\textwidth]{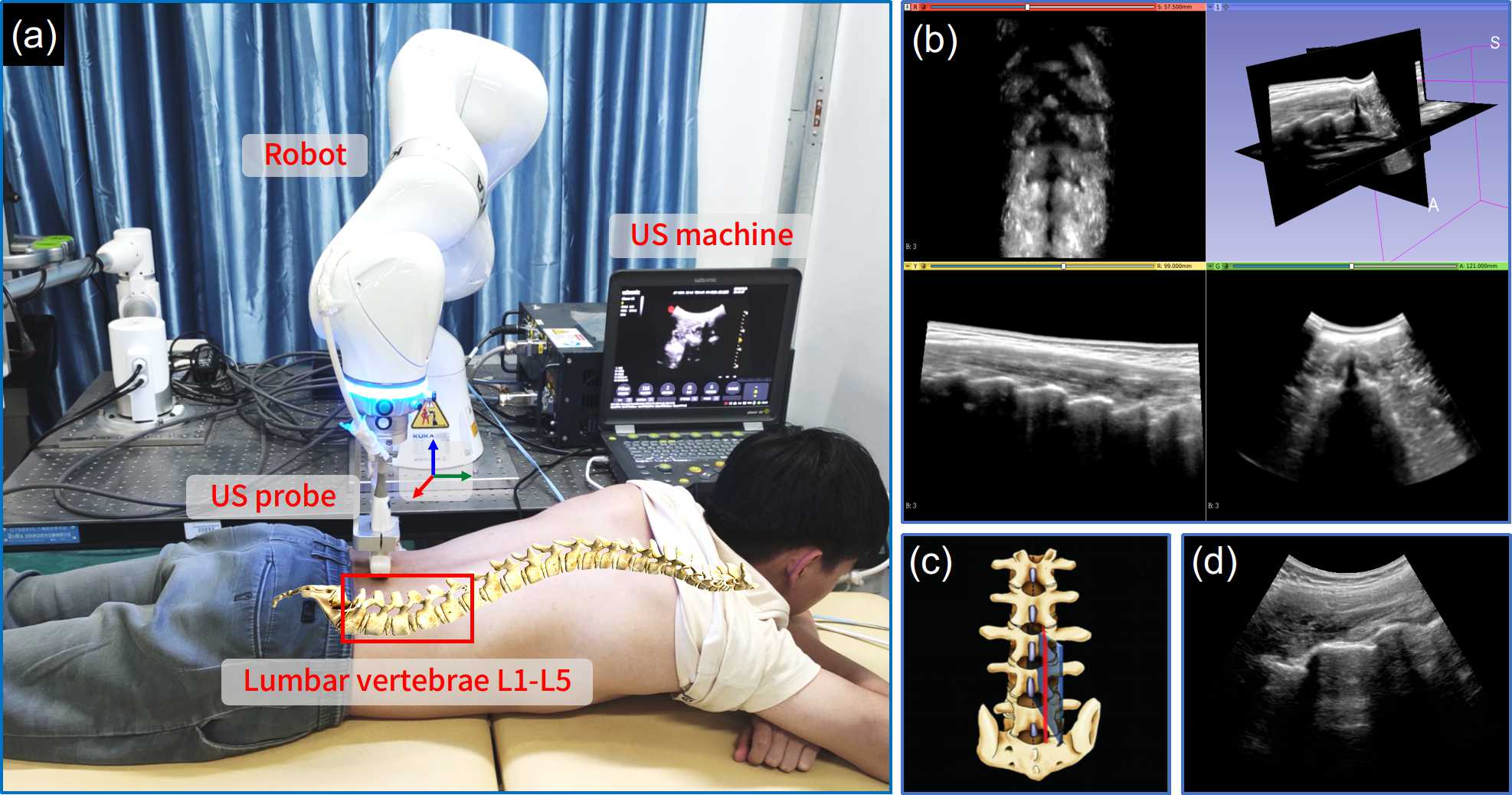}
\caption{Illustration of the data acquisition. (a) A robotic US system is used to acquire B-mode US images of the L1-L5 lumbar vertebrae of the volunteers for 3D volume reconstruction. (b) is the visualization of an exemplary data volume in 3D Slicer \cite{slicer}. (c) illustrates the \textit{paramedian sagittal lamina view} (PSL) of the spine \cite{spineUS}, and (d) shows an image of the PSL plane acquired by a clinician.}
\label{Fig_data_acquisition}
\vspace{-0.5cm}
\end{figure}

\section{EXPERIMENTS}

In order to demonstrate the effectiveness of our proposed RL framework, we apply it to the task of US imaging of the spine, to autonomously navigate the probe towards one of the standard scan planes of the spine -- the \textit{paramedian sagittal lamina view} (PSL) \cite{spineUS}, as illustrated in Fig. \ref{Fig_data_acquisition} (c)(d).

\subsection{Data Acquisition}
A total of $41$ 3D US volumes of the spine that cover the L1-L5 lumbar vertebrae are acquired from $17$ healthy male volunteers aged $20$ to $26$. The average volume size of our dataset is $350\times397\times274$ and each voxel is $0.5\times0.5\times0.5mm^3$. In order to acquire the data volumes, we built a robotic US system using a KUKA LBR iiwa 7 R800 (KUKA Roboter GmbH, Augsburg, Germany) with a C5-1B convex US transducer mounted at its end-effector and connected to a Wisonic Clover diagnostic US machine (Shenzhen Wisonic Medical Technology Co., Ltd, China), as shown in Fig. \ref{Fig_data_acquisition} (a). The volunteers were in a prone position on an examination bed during the scan.
Before acquisition, a clinician selected the scanning parameters, applied the coupling gel on the surface of the volunteers and specified the start and end points of the scan. During acquisition, the robot linearly moved the probe from the start point to the end point under Cartesian impedance control. A high stiffness value ($2000 N/m$) was set in the $xy$ plane and a low stiffness value ($50 N/m$) was set along the $z$-axis, and an additional force of $5 N$ was applied in the downward direction to ensure contact between the probe and the patient. The B-mode images and the probe pose measured by the robot were transmitted to a computer for volume reconstruction using a squared distance weighted approach \cite{huang2005development}. An exemplary data volume is shown in Fig. \ref{Fig_data_acquisition} (b). In addition, the clinician manually navigated the probe towards the PSL plane (see Fig. \ref{Fig_data_acquisition} (d)) and the corresponding probe pose was recorded.

\begin{table*}[hbt] \renewcommand\arraystretch{1.1} \small
\centering
\setlength{\abovecaptionskip}{0.05cm}  
\caption{Quantitative Results of US Probe Navigation Towards the Standard Scan Plane}
\begin{tabular}{|l|l||c|c|c|c|c|c|c|}
\hline
\multicolumn{2}{|c||}{\multirow{3}{*}{\tabincell{l}{Methods}}}& \multirow{3}{*}{\tabincell{l}{$\Delta d +\Delta \theta$}} & \multirow{3}{*}{\tabincell{c}{Position error ($mm$)}} & \multirow{3}{*}{\tabincell{c}{Orientation error ($^\circ$)}} & \multirow{3}{*}{\tabincell{l}{SSIM}} & \multirow{3}{*}{\tabincell{c}{Success \\ rate}} &\multirow{3}{*}{\tabincell{c}{Average \\number \\of steps}} \\ 
\multicolumn{2}{|c||}{} & & & & & & \\
\multicolumn{2}{|c||}{} & & & & & & \\
\hline
\multicolumn{2}{|l||}{Intra-observer errors}& --& $4.30 \pm 1.53$& $ 3.34 \pm1.60$ & $ 0.68 \pm0.14$& --&--\\
\hline
\multirow{2}{*}{\tabincell{l}{Intra- \\ patient}} & SonoRL w/o conf &$0.34 \pm 0.15$&$5.85\pm6.77$&$13.00\pm32.20$& $0.66 \pm 0.18$ & $79\% $ & $61$\\
\cline{2-8}
& SonoRL w/ conf &$\mathbf{0.35 \pm 0.13}$&$\mathbf{4.91\pm4.44}$&$\mathbf{4.65\pm2.61}$&$\mathbf{0.67 \pm 0.21}$ &$ \mathbf{92\%}$ &$66$\\
  \hline
\multirow{2}{*}{\tabincell{l}{Inter- \\patient}} & SonoRL w/o conf &$0.16\pm 0.17 $&$21.87\pm15.30$&$39.93\pm61.04$&$0.43 \pm 0.16$ & $21\% $ &$63$ \\ 
  \cline{2-8}
  & SonoRL w/ conf & $\mathbf{0.18 \pm 0.17}$&$\mathbf{13.93\pm11.58}$&$\mathbf{29.12\pm51.46}$&$\mathbf{0.50 \pm 0.23}$  & $\mathbf{46\%} $ &$76$ \\ 
  \hline
  \end{tabular}
  \label{quantitative}
\vspace{-0.4cm}
\end{table*}

\subsection{Simulation Setup}
We built a simulation environment in Python for US probe navigation with the \textit{SonoRL} algorithm. At each time step, the agent observes an image of size $150 \times 150$ and stacks $4$ recent frames as the state. The ROI is manually selected with a size of $110\times90$. In each episode, the horizontal position of the probe is randomly initialized in the center region $\{(x,y):x \sim \mathcal{U}(0.3W,0.7W),y\sim \mathcal{U}(0.2L,0.8L)\}$, where $L$, $W$ are the length and width of the data volume, and the initial $z$-coordinate of the probe is determined by the extracted surface. The initial $z$-axis of the probe is aligned with the $-z$ direction of the world frame, and the probe is randomly rotated $\eta\sim \mathcal{U}(0,360^\circ)\}$ around its $z$-axis. The maximum number of steps in each episode is limited to $120$.

\subsection{Quantitative Evaluation}

\begin{figure}[bt]
\setlength{\abovecaptionskip}{0.1cm}  
\centering
\includegraphics[scale=1.0,angle=0,width=0.48\textwidth]{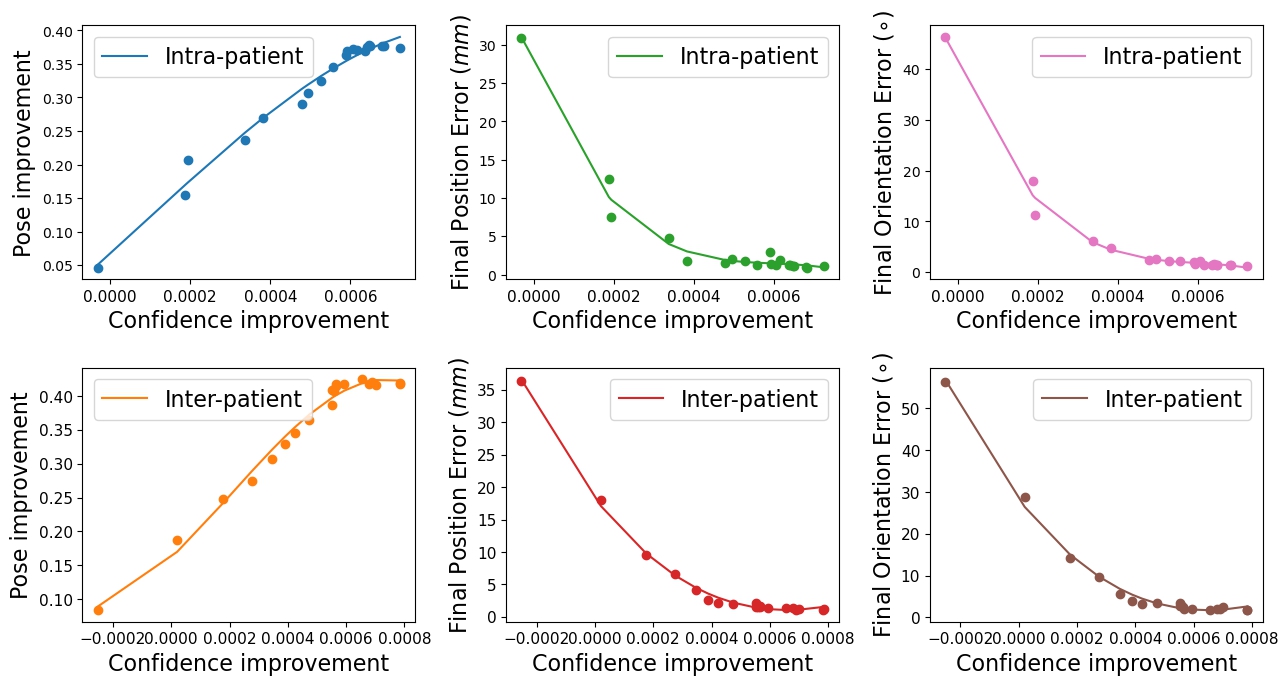}
\caption{The navigation performance against the image quality improvement of the \textit{SonoRL w/o conf} agent during training in the intra- and inter-patient settings. As the average confidence improvement per step increases, the average pose improvement per step increases and the final position and orientation errors decrease, showing that the navigation performance is positively correlated with the improvement of image quality.}
\label{Fig_training_metrics_vs_confidence}
\vspace{-0.5cm}
\end{figure}

In order to fully evaluate the effectiveness of our method, we consider two different settings: \textit{intra-patient} and \textit{inter-patient}. In the intra-patient setting, the model is trained with $33$ data volumes obtained from $17$ subjects and tested with $8$ unseen data volumes obtained from $8$ of these subjects. The design of this setting is motivated by the real-world scenario where more than one US acquisition of the same patient is required, such as pre- and post-operative ultrasonography. In the inter-patient setting, the model is trained with $33$ data volumes obtained from $14$ subjects and tested with $8$ data volumes obtained from $3$ unseen subjects. This task is more difficult since it requires the learned policy to be generalized to patients with highly variable anatomical structures. The \textit{SonoRL} agents with and without confidence optimization are refered to as \textit{SonoRL w/ conf} and \textit{SonoRL w/o conf}, respectively. Each model is optimized for $160k$ and $200k$ iterations in the intra- and inter-patient settings respectively to achieve stable performance.

We first studied the relationship between the \textit{SonoRL w/o conf} agent's navigation performance and the improvement of image quality during training, as shown in Fig. \ref{Fig_training_metrics_vs_confidence}. In both the intra- and inter-patient settings, as the confidence improvement in each step becomes greater, the pose improvement in each step increases and the final pose error decreases. This indicates that the navigation performance is positively correlated with the improvement of image quality, as predicted in Section~III.B. Therefore, although only $\Delta d +\Delta \theta$ is used in the reward function (\ref{reward}), the agent also implicitly learns to increase $\Delta c$ as it gradually learns to correctly navigate the probe.

The quantitative evaluation of the two agents is carried out on $24$ random test cases for each setting, including $3$ navigation tests on each virtual patient. We compare the performance of the two agents and the intra-observer errors of a human expert in Table \ref{quantitative}. The metrics include the average pose improvement per step ($\Delta d +\Delta \theta$), the final pose errors, the structural similarity (SSIM) \cite{SSIM} between the final plane image and the goal image, success rate and the average number of steps. The navigation is considered successful if the final pose error is less than $10mm/10^\circ$.

In the intra-patient setting, the \textit{SonoRL} agents can successfully accomplish most of the tasks, and the final pose error of the \textit{SonoRL w/ conf} agent is similar to intra-observer errors, indicating that our method is able to perform reproducible probe navigation on familiar patients. In both settings, as the \textit{SonoRL w/ conf} agent tends to avoid locations with poor image quality, it takes more navigation steps than the \textit{SonoRL w/o conf} agent, but achieves higher pose improvement and has a greater chance of successfully reaching the goal. Both agents show a degraded performance in the inter-patient setting, mainly because the task is more difficult and challenging. Nevertheless, the introduction of confidence optimization improves the navigation performance by a large margin, which indicates that the optimization of image quality improves the generalization of the learned policy to highly variable patient anatomy. 

\begin{figure*}[hbt]
\setlength{\abovecaptionskip}{0.1cm}  
\centering
\includegraphics[scale=1.0,angle=0,width=0.99\textwidth]{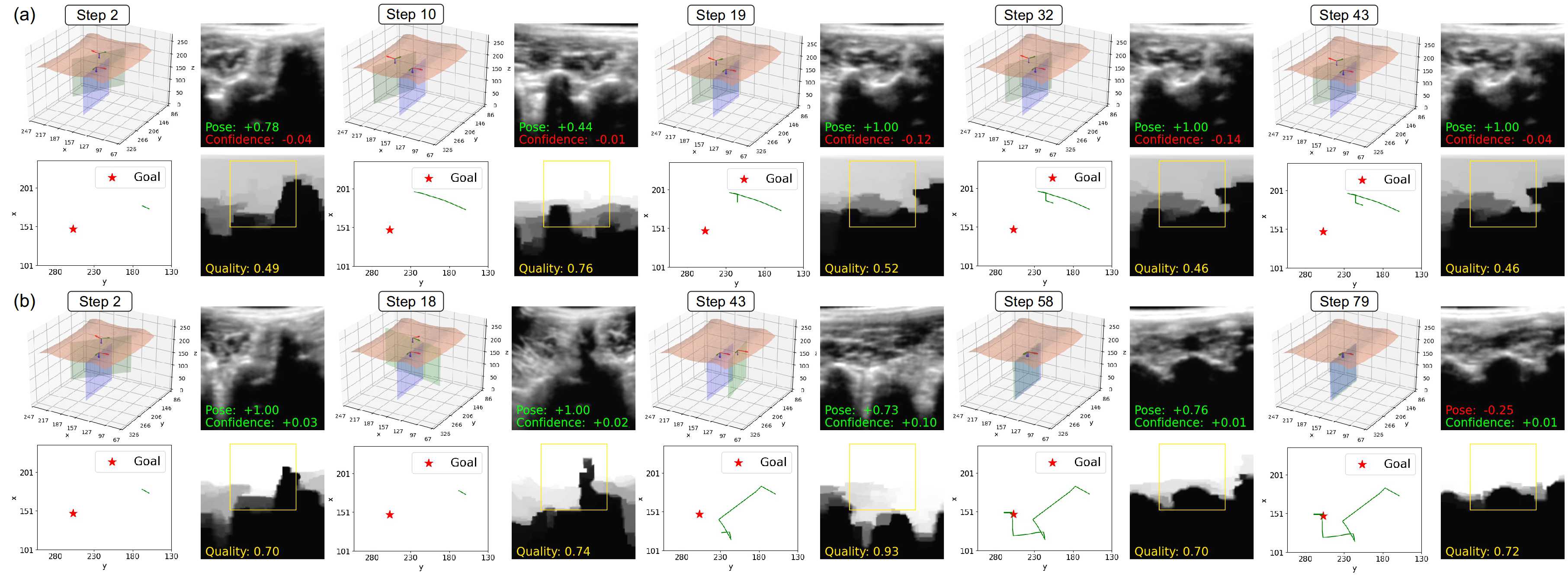}
\caption{Snapshots of the trajectories of the (a) \textit{SonoRL w/o conf} and (b) \textit{SonoRL w/ conf} agents in an intra-patient test case. The 3D plot shows the current plane (green), the goal plane (blue), the patient surface (salmon) and the poses of the current probe and the goal. Below it shows the top-view trajectory of the agent on the $xy$ plane (green), and the goal position is indicated by a red star. The current plane image is displayed on the right side of each 3D plot, with the pose and confidence improvement of the current step marked in green (positive) or red (negative). The confidence map below it shows the pixel-wise confidence in the current image with the average confidence in the ROI (yellow rectangle).}
\label{Fig_traj}
\vspace{-0.4cm}  
\end{figure*}

\subsection{Qualitative Evaluation}

We further compare our methods with and without confidence optimization through qualitative analysis of an intra-patient test case, where the \textit{SonoRL w/o conf} agent fails to navigate to the goal but the \textit{SonoRL w/ conf} agent successfully reaches the goal with an accuracy of $1.04 mm/2.18 ^\circ$. The trajectories of the two agents are visualized in Fig. \ref{Fig_traj}.

As shown in Fig. \ref{Fig_traj} (a), the \textit{SonoRL w/o conf} agent greatly reduces the position error in the first $10$ steps. However, since the agent is not aware of the degradation of image quality, it blindly chooses actions to reduce the pose error and navigates to a location with poor image quality (step $19$, $c=0.52$). In the remaining steps, the agent is stuck in this place until the end of the navigation (step $43$) and fails to reach the goal. 

In contrast, the \textit{SonoRL w/ conf} agent adjusts its orientation in the first $18$ steps to improve both the pose and the image quality. Then, the agent gradually approaches the goal in steps $19$ to $58$ based on the high-quality images, and further fine-tunes its pose in the remaining $21$ steps. Since the confidence-aware agent considers both the distance-to-goal and the image quality when making decisions, it may not choose the most aggresive actions to achieve the goal, but will carefully avoid the bad acoustic windows during the navigation. Therefore, it ends up with a relatively longer navigation time and the trajectory appears circuitous (see Fig. \ref{Fig_traj} (b)). However, this strategy allows the agent to approach the goal more stably because the acquired images are clear and contain key information of the anatomical structures to guide the navigation. 

\section{CONCLUSIONS}

In this paper, we propose a deep RL framework to navigate a virtual US probe towards the standard scan planes, taking into account several practical requirements in real-world US scans. The optimization of image quality is encoded in the learning process through a confidence-based method to improve the navigation performance. As a next step, the methods presented in this paper will be integrated with existing robotic US systems for real-world applications, where additional challenges may need to be tackled. In addition, we plan to develop a soft end-effector to hold the probe and investigate novel actuation methods such as soft fluidic actuation \cite{lindenroth2019design} and magnetic actuation \cite{xu2020novel}\cite{xu2020improved} for the probe control task to improve the compliance of the system.

\addtolength{\textheight}{-1cm}   




%
%

\bibliographystyle{IEEEtran}   
\bibliography{root}

\begin{thebibliography}{10}
\providecommand{\url}[1]{#1}
\csname url@samestyle\endcsname
\providecommand{\newblock}{\relax}
\providecommand{\bibinfo}[2]{#2}
\providecommand{\BIBentrySTDinterwordspacing}{\spaceskip=0pt\relax}
\providecommand{\BIBentryALTinterwordstretchfactor}{4}
\providecommand{\BIBentryALTinterwordspacing}{\spaceskip=\fontdimen2\font plus
\BIBentryALTinterwordstretchfactor\fontdimen3\font minus
  \fontdimen4\font\relax}
\providecommand{\BIBforeignlanguage}[2]{{%
\expandafter\ifx\csname l@#1\endcsname\relax
\typeout{** WARNING: IEEEtran.bst: No hyphenation pattern has been}%
\typeout{** loaded for the language `#1'. Using the pattern for}%
\typeout{** the default language instead.}%
\else
\language=\csname l@#1\endcsname
\fi
#2}}
\providecommand{\BIBdecl}{\relax}
\BIBdecl

\bibitem{ultrasound}
K.~K. Shung, ``Diagnostic ultrasound: Past, present, and future,'' \emph{J Med
  Biol Eng}, vol.~31, no.~6, pp. 371--4, 2011.

\bibitem{disorder1}
G.~Brown, ``Work related musculoskeletal disorders in sonographers,''
  \emph{BMUS Bulletin}, vol.~11, no.~3, pp. 6--13, 2003.

\bibitem{disorder2}
M.~Muir, P.~Hrynkow, R.~Chase, D.~Boyce, and D.~Mclean, ``The nature, cause,
  and extent of occupational musculoskeletal injuries among sonographers:
  recommendations for treatment and prevention,'' \emph{Journal of Diagnostic
  Medical Sonography}, vol.~20, no.~5, pp. 317--325, 2004.

\bibitem{mustafa2013robio}
A.~S.~B. Mustafa, T.~Ishii, Y.~Matsunaga, R.~Nakadate, H.~Ishii, K.~Ogawa,
  A.~Saito, M.~Sugawara, K.~Niki, and A.~Takanishi, ``Development of robotic
  system for autonomous liver screening using ultrasound scanning device,'' in
  \emph{2013 IEEE International Conference on Robotics and Biomimetics
  (ROBIO)}.\hskip 1em plus 0.5em minus 0.4em\relax IEEE, 2013, pp. 804--809.

\bibitem{Hennersperger2016MRI}
C.~Hennersperger, B.~Fuerst, S.~Virga, O.~Zettinig, B.~Frisch, T.~Neff, and
  N.~Navab, ``Towards mri-based autonomous robotic us acquisitions: a first
  feasibility study,'' \emph{IEEE transactions on medical imaging}, vol.~36,
  no.~2, pp. 538--548, 2016.

\bibitem{virga2016iros}
S.~Virga, O.~Zettinig, M.~Esposito, K.~Pfister, B.~Frisch, T.~Neff, N.~Navab,
  and C.~Hennersperger, ``Automatic force-compliant robotic ultrasound
  screening of abdominal aortic aneurysms,'' in \emph{2016 IEEE/RSJ
  International Conference on Intelligent Robots and Systems (IROS)}.\hskip 1em
  plus 0.5em minus 0.4em\relax IEEE, 2016, pp. 508--513.

\bibitem{huang2018robotic}
Q.~Huang, J.~Lan, and X.~Li, ``Robotic arm based automatic ultrasound scanning
  for three-dimensional imaging,'' \emph{IEEE Transactions on Industrial
  Informatics}, vol.~15, no.~2, pp. 1173--1182, 2018.

\bibitem{baumgartner2017sononet}
C.~F. Baumgartner, K.~Kamnitsas, J.~Matthew, T.~P. Fletcher, S.~Smith, L.~M.
  Koch, B.~Kainz, and D.~Rueckert, ``Sononet: real-time detection and
  localisation of fetal standard scan planes in freehand ultrasound,''
  \emph{IEEE transactions on medical imaging}, vol.~36, no.~11, pp. 2204--2215,
  2017.

\bibitem{spineUS}
\BIBentryALTinterwordspacing
M.~K. Karmakar and K.~J. Chin, \emph{Spinal Sonography and Applications of
  Ultrasound for Central Neuraxial Blocks}.\hskip 1em plus 0.5em minus
  0.4em\relax New York, NY: McGraw-Hill Education, 2017. [Online]. Available:
  \url{accessanesthesiology.mhmedical.com/content.aspx?aid=1141735352}
\BIBentrySTDinterwordspacing

\bibitem{li2019sarl}
K.~Li, Y.~Xu, J.~Wang, and M.~Q.-H. Meng, ``{SARL$^*$}: Deep reinforcement
  learning based human-aware navigation for mobile robot in indoor
  environments,'' in \emph{2019 IEEE International Conference on Robotics and
  Biomimetics (ROBIO)}.\hskip 1em plus 0.5em minus 0.4em\relax IEEE, 2019, pp.
  688--694.

\bibitem{alansary2018automatic}
A.~Alansary, L.~Le~Folgoc, G.~Vaillant, O.~Oktay, Y.~Li, W.~Bai,
  J.~Passerat-Palmbach, R.~Guerrero, K.~Kamnitsas, B.~Hou \emph{et~al.},
  ``Automatic view planning with multi-scale deep reinforcement learning
  agents,'' in \emph{International Conference on Medical Image Computing and
  Computer-Assisted Intervention}.\hskip 1em plus 0.5em minus 0.4em\relax
  Springer, 2018, pp. 277--285.

\bibitem{alansary2019evaluating}
A.~Alansary, O.~Oktay, Y.~Li, L.~Le~Folgoc, B.~Hou, G.~Vaillant, K.~Kamnitsas,
  A.~Vlontzos, B.~Glocker, B.~Kainz \emph{et~al.}, ``Evaluating reinforcement
  learning agents for anatomical landmark detection,'' \emph{Medical image
  analysis}, vol.~53, pp. 156--164, 2019.

\bibitem{confimap}
A.~Karamalis, W.~Wein, T.~Klein, and N.~Navab, ``Ultrasound confidence maps
  using random walks,'' \emph{Medical image analysis}, vol.~16, no.~6, pp.
  1101--1112, 2012.

\bibitem{schmidt2019offset}
A.~Schmidt-Richberg, N.~Schadewaldt, T.~Klinder, M.~Lenga, R.~Trahms,
  E.~Canfield, D.~Roundhill, and C.~Lorenz, ``Offset regression networks for
  view plane estimation in 3d fetal ultrasound,'' in \emph{Medical Imaging
  2019: Image Processing}, vol. 10949.\hskip 1em plus 0.5em minus 0.4em\relax
  International Society for Optics and Photonics, 2019, p. 109493K.

\bibitem{li2018standard}
Y.~Li, B.~Khanal, B.~Hou, A.~Alansary, J.~J. Cerrolaza, M.~Sinclair,
  J.~Matthew, C.~Gupta, C.~Knight, B.~Kainz \emph{et~al.}, ``Standard plane
  detection in 3d fetal ultrasound using an iterative transformation network,''
  in \emph{International Conference on Medical Image Computing and
  Computer-Assisted Intervention}.\hskip 1em plus 0.5em minus 0.4em\relax
  Springer, 2018, pp. 392--400.

\bibitem{dou2019agent}
H.~Dou, X.~Yang, J.~Qian, W.~Xue, H.~Qin, X.~Wang, L.~Yu, S.~Wang, Y.~Xiong,
  P.-A. Heng \emph{et~al.}, ``Agent with warm start and active termination for
  plane localization in 3d ultrasound,'' in \emph{International Conference on
  Medical Image Computing and Computer-Assisted Intervention}.\hskip 1em plus
  0.5em minus 0.4em\relax Springer, 2019, pp. 290--298.

\bibitem{burke2020learning}
M.~Burke, K.~Lu, D.~Angelov, A.~Strai{\v{z}}ys, C.~Innes, K.~Subr, and
  S.~Ramamoorthy, ``Learning robotic ultrasound scanning using probabilistic
  temporal ranking,'' \emph{arXiv preprint arXiv:2002.01240}, 2020.

\bibitem{droste2020automatic}
R.~Droste, L.~Drukker, A.~T. Papageorghiou, and J.~A. Noble, ``Automatic probe
  movement guidance for freehand obstetric ultrasound,'' in \emph{International
  Conference on Medical Image Computing and Computer-Assisted
  Intervention}.\hskip 1em plus 0.5em minus 0.4em\relax Springer, 2020, pp.
  583--592.

\bibitem{jarosik2019automatic}
P.~Jarosik and M.~Lewandowski, ``Automatic ultrasound guidance based on deep
  reinforcement learning,'' in \emph{2019 IEEE International Ultrasonics
  Symposium (IUS)}.\hskip 1em plus 0.5em minus 0.4em\relax IEEE, 2019, pp.
  475--478.

\bibitem{milletari2019straight}
F.~Milletari, V.~Birodkar, and M.~Sofka, ``Straight to the point: reinforcement
  learning for user guidance in ultrasound,'' in \emph{Smart Ultrasound Imaging
  and Perinatal, Preterm and Paediatric Image Analysis}.\hskip 1em plus 0.5em
  minus 0.4em\relax Springer, 2019, pp. 3--10.

\bibitem{hase2020ultrasound}
H.~Hase, M.~F. Azampour, M.~Tirindelli, M.~Paschali, W.~Simson, E.~Fatemizadeh,
  and N.~Navab, ``Ultrasound-guided robotic navigation with deep reinforcement
  learning,'' \emph{arXiv preprint arXiv:2003.13321}, 2020.

\bibitem{mnih2015human}
V.~Mnih, K.~Kavukcuoglu, D.~Silver, A.~A. Rusu, J.~Veness, M.~G. Bellemare,
  A.~Graves, M.~Riedmiller, A.~K. Fidjeland, G.~Ostrovski \emph{et~al.},
  ``Human-level control through deep reinforcement learning,'' \emph{nature},
  vol. 518, no. 7540, pp. 529--533, 2015.

\bibitem{2017confidriven}
P.~Chatelain, A.~Krupa, and N.~Navab, ``Confidence-driven control of an
  ultrasound probe,'' \emph{IEEE Transactions on Robotics}, vol.~33, no.~6, pp.
  1410--1424, 2017.

\bibitem{kingma2014adam}
D.~P. Kingma and J.~Ba, ``Adam: A method for stochastic optimization,''
  \emph{arXiv preprint arXiv:1412.6980}, 2014.

\bibitem{slicer}
A.~Fedorov, R.~Beichel, J.~Kalpathy-Cramer, J.~Finet, J.-C. Fillion-Robin,
  S.~Pujol, C.~Bauer, D.~Jennings, F.~Fennessy, M.~Sonka \emph{et~al.}, ``3d
  slicer as an image computing platform for the quantitative imaging network,''
  \emph{Magnetic resonance imaging}, vol.~30, no.~9, pp. 1323--1341, 2012.

\bibitem{huang2005development}
Q.-H. Huang, Y.-P. Zheng, M.-H. Lu, and Z.~Chi, ``Development of a portable 3d
  ultrasound imaging system for musculoskeletal tissues,'' \emph{Ultrasonics},
  vol.~43, no.~3, pp. 153--163, 2005.

\bibitem{SSIM}
Z.~Wang, A.~C. Bovik, H.~R. Sheikh, and E.~P. Simoncelli, ``Image quality
  assessment: from error visibility to structural similarity,'' \emph{IEEE
  transactions on image processing}, vol.~13, no.~4, pp. 600--612, 2004.

\bibitem{lindenroth2019design}
L.~Lindenroth, R.~J. Housden, S.~Wang, J.~Back, K.~Rhode, and H.~Liu, ``Design
  and integration of a parallel, soft robotic end-effector for extracorporeal
  ultrasound,'' \emph{IEEE Transactions on Biomedical Engineering}, 2019.

\bibitem{xu2020novel}
Y.~Xu, K.~Li, Z.~Zhao, and M.~Q.-H. Meng, ``A novel system for closed-loop
  simultaneous magnetic actuation and localization of wce based on external
  sensors and rotating actuation,'' \emph{IEEE Transactions on Automation
  Science and Engineering}, 2020.

\bibitem{xu2020improved}
{Y. Xu}, {K. Li}, {Z. Zhao}, and {M. Q.-H. Meng}, ``Improved multiple objects
  tracking based autonomous simultaneous magnetic actuation \& localization for
  wce,'' in \emph{2020 IEEE International Conference on Robotics and Automation
  (ICRA)}.\hskip 1em plus 0.5em minus 0.4em\relax IEEE, 2020, pp. 5523--5529.

\end{thebibliography}

\end{document}